# Emotion Detection using Data Driven Models

Naveenkumar K S, Vinayakumar R, Soman KP

Center for Computational Engineering and Networking (CEN),Amrita School of Engineering, Coimbatore, Amrita Vishwa Vidyapeetham, India
Email: naveensivakumarr@gmail.com

**Abstract**

Text is the major method that is used for communication now a days, each and every day lots of text are created. In this paper the text data is used for the classification of the emotions. Emotions are the way of expression of the person's feelings which has an high influence on the decision making tasks. Datasets are collected which are available publically and combined together based on the three emotions that are considered here positive, negative and neutral. In this paper we have proposed the text representation method TFIDF and keras embedding and then given to the classical machine learning algorithms of which Logistics Regression gives the highest accuracy of about 75.6%, after which it is passed to the deep learning algorithm which is the CNN which gives the state of art accuracy of about 45.25%. For the research purpose the datasets that has been collected are released.

## 1 Introduction

Emotions are the combinations of the feelings, behavior, physiology, conceptualization and experience [1] [2] that are expressed by any living beings. The emotions can be expressed through facial expressions, text representation and through speech. Here in Natural language processing we consider the text as the major method for the expression of the emotion. The text depends on the language that is used for communicating and the language contains the emotion and the intensity of the emotion in which we are using them [3]. Emotions are the one which all are bound to it every action that we do results our emotion. Text is the way were majority of the people communicate with each other which contains a lot of the information about the persons state of mind. Now a days we are living in the world were all the information are shared in the internet were all the individuals are having easy access to it. Each human has their identity in the digital world through the social medias such as the Facebook, twitter, Instagram, gmail, snapchat, whatsup and what so ever application the user is bound to use it. Tons of the text are generated each and every day in these social media. Peoples use these common social media to express their feelings nothing but the emotion towards anything or anybody they may be personal or may be a social impact based on the individual. Emotions that are considered here are basically positive, negative and neutral. The datasets which are annotated and available are collected and combined together resulting in the creation of the new dataset based on the existing datasets. The task here is to identify and to differentiate the text that is given into these three classes when they are tested with the test dataset. Here in multiclass problem text representation methods are used such as TFIDF and keras embedding [13-22] which is then passed to classical machine learning algorithm and then the deep learning algorithms for the purpose of the classification. These methods have performed well in various tasks in recent days in compared to classical machine learning classifiers [23] [24] [25] [26]. The objective of this paper is to create a training model which is basically an intelligent system which understands the type of the emotion based on the texts which are generated. The paper contains the following parts such as the Related works, Description of the dataset, Background, Experimentation and observations, Conclusion and References.

## 2 Related works

It is a shared task done by detecting the intensity of the emotion that has been expressed by the speaker of that tweet. The author of this paper has said that the dataset has been created by using the two methods such as the worst scaling method and the crowd sourcing method.[4] From this paper the author says that a standard regression system had been created and various experiments have been conducted to show the affected lexicons which shows the scores that are related to the word emotion that are helpful in determining the emotion intensity. The datasets has been collected and annotated manually and then various machine learning and deep learning approaches are being used which included the keras embedding and LSTM.[3] [5] The author of this paper deals with the unsupervised learning which mainly focus on the context based approach for the detection of the emotions in the text in a sentence level.[6] From this paper the understanding is made clear by the author about the syntactic dependency parsing and this method have been used in the paper for different  approach to extract the similar words that are related  to the affected word syntactic dependencies. This paper talks about the information-theoretic interpretation of the TFIDF and has given a view of the probability of the occurrence of the information that are present in a term weighted bytes [7]. The author deals with the probability weighted amount of the information which are shortly called as the PEI that are useful in measuring the terms in the documents that are based on the information-theoretic view of retrieval events [10]. This paper deals with the polarity classification and quantification of the English tweets which involves in the taking of the information from the message level and the topic based sentiment [8]. LSTM has been used in this method with the two kinds of the attention mechanisms [9].

## 3 Description of the Dataset

The dataset used in this paper is the twitter datasets which are annotated and readily available publicly. The collected dataset are preprocessed and then extracted the feature using the text representation methods and given to the machine learning and the deep learning algorithms. At first the datasets which are annotated and collected and then they combined in a random manner together based on the three emotions such as the positive negative and neutral. The datasets are split into the train and the test as the 70% and 30% respectively. The detailed description of the dataset are given below. The dataset consist of totally 131601 instances the split and the total number of the data are given below.

The datasets that are formed are visualized using the Tagcrowd software which is for the visualization of the text data. Tags are the single words which are shown by the different color by showing the importance of the each data with the different font and color. There are the certain parameters that are to be set as a prerequisites such as the Maximum number of the words to show, Minimum frequency of the words, Group similar frequency, Lower case and don't show words after filling this the data is loaded and then visualized. The three emotion dataset are loaded and the corresponding results are being shown below:

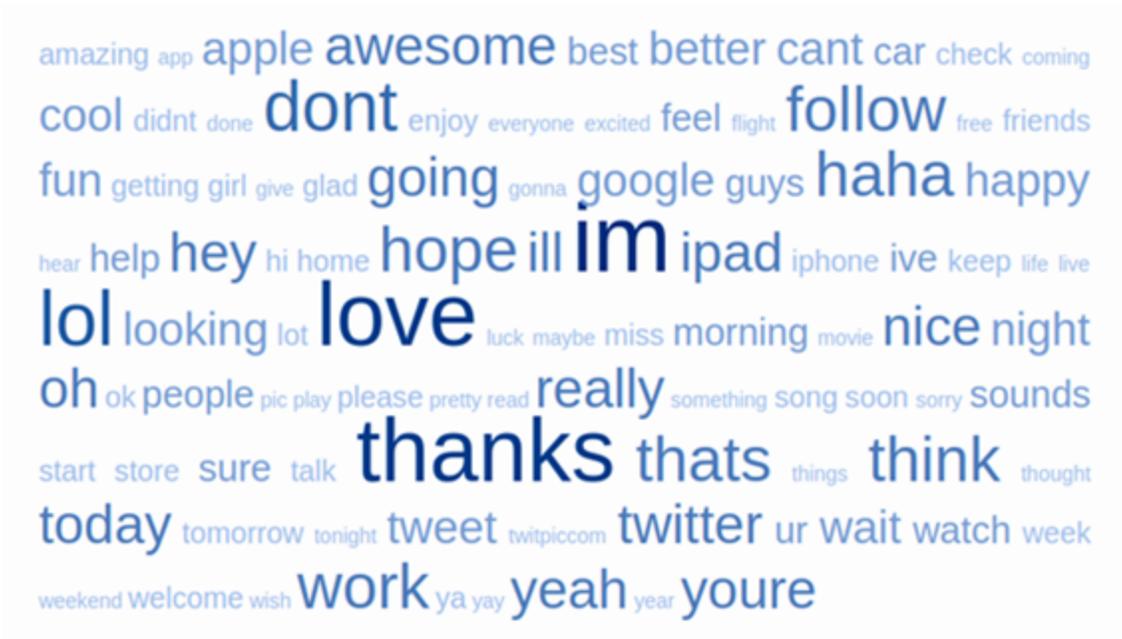

**Fig 1 Plot for Positive Dataset**

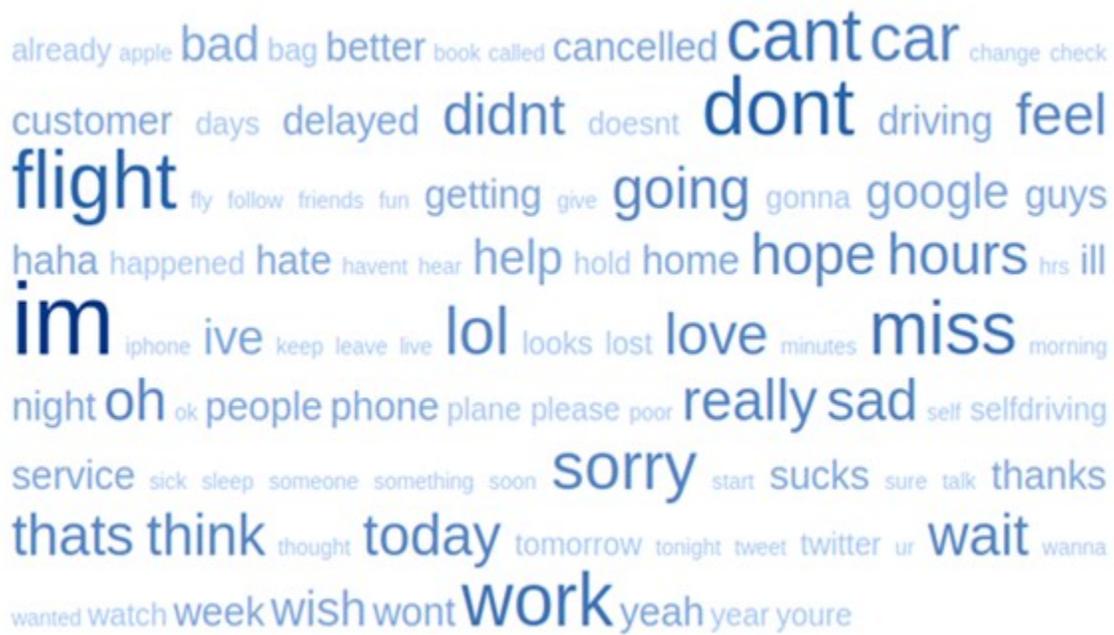

**Fig 2 Plot for Negative Dataset**

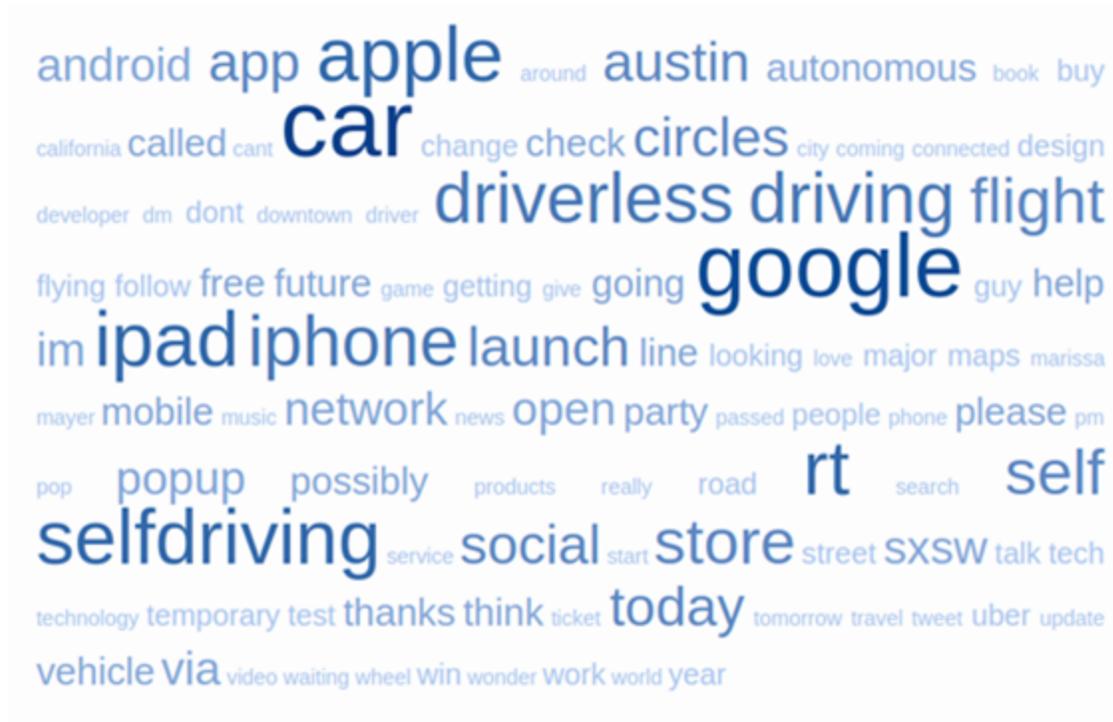

**Fig 3 Plot for Neutral Dataset**

| Category | Sentences |
|---|---|
| Train | 92120 |
| Test | 39481 |
| Positive | 62629 |
| Negative | 55477 |
| Neutral | 13495 |

**Table 1 Description of Data set**

### 4.1 Term Frequency Inverse Document Frequency (TFIDF)

A Vectorizer that performs the task of the conversion of the text to the feature by giving weights often used for information extraction and is commonly called as the Term Frequency Inverse Document Frequency .[5] The TFIDF gives the information about how much that word is important to that document with resepect to the corpus. The tfidf works in such a way that it calculates the value based on the weights that are the central tool for finding the rank for the document. They work in such a way that each and every sentence in the document is converted into the vector.[9] The tfidf has the two parts as the name indicates they are the term frequency and inverse document frequency [5]. The function of the term frequency is that it finds out the number of times the word occurs in the document but the inverse document frequency works in a different manner such as they are calculated by taking the log of the number of the documents present in the corpus divided by the number of the documents in which this terms appears.[9]. The feature can be extracted by using mindf and maxdf which are nothing but the maximum document frequency and minimum document frequency. The maxdf is a corpus specific and used for the removal of the words that are occurring too frequently, in contrary to it is the mindf are used for the checking of the words that do not occur in at least n documents were the value of the mindf = n.

The mathematical representation of the IDF is given below as:

IDF(t) = log[{D/DF(t)]]

where D is the total number of the documents

W_i = TF(t_i,d).IDF(t_i)

Where Wi is called as the weight of the word Ti in the document D

### 4.2 Keras Embedding

Embedding is a common technique which is used explicitly in natural language processing for the word level extraction of features which are nothing but the information from the particular context [3]. They provide a dense representation of the words and their relative meanings which is an improved model over the sparse representations that are being used in the bag of words representation. When it comes to the keras embedding the data which is given as the input must have the integer encoded so that each word is represented as the unique integer [10] . The Embedding layer is set with the random weights that will learn an embedding for all of the words in the training dataset.

### 4.3 Logistic Regression

The concept of the logistics regression has a much more borrowed concepts form the statistics to interpret them easily were there is small difference in them with that of the linear regression were it prevents the misclassification in the case of the linear regression there are independence between the features. This method is a predictive analysis similar to all other regression methods. They are used to find the relationship between the dependent and the independent variables such as the binary variable and the one or more ordinal and normal or ratio level.[10] The logistics regression has one term called as the maximum likelihood function they work in a method such as instead of minimizing the squared residuals for finding the best fitting line this function can be used.[11] The Maximum likelihood function commonly called as the ML is used for finding the shortest possible deviance between the predicted values and the observed values for finding the best fitting line using the calculus. This can be achieved by using different iterations performed by the computer until the shortest possible deviance is found for the best fit. [7].

### 4.4 Convolution Neural Network (CNN)

CNN which is commonly called as the Convolution Neural Network is used which works on the principle of the method of the convolution. The convolution which is being used here is the 1-D convolution which means the dataset which is collected are vectorised using the keras embedding method and then they are passed to the 1-D convolution which means the 3x3 filter are used for the process of the convolving and then the maxpooled output is being taken and a matrix is being formed based on the output of the filter.[12] Then the convolved output forms the new input to the other layer is so it is present.

### 5 Experiments and Observations

The experiment is conducted in the lab with the computation power of the processor. The dataset which is collected are preprocessed and then the dataset is passed to the text representation method TFIDF were the feature extraction rate is varied by keeping it from 10000 to 40000 various trail experiments were conducted in order to achieve this state of art accuracy were when the features are in the low order the logistics regression outperforms well and as the features are increased the support vector machine that too specifically the linear kernel outperforms well. Then the other classical machine learning algorithms are used taking into account the scikit learn library other classifiers such as the Decision tree, Adaboost, RandomForest tree, SVM kernel rbf are used. The results that are obtained are done a comparative study by drawing a table with the results that are got. After passing it to the classical machine learning algorithms the dataset is passed to the deep learning algorithm after giving it to the keras embedding technique which is done for the word level feature extraction and then given to the Convolution Neural Network which is commonly called as the CNN the results that are obtained by using this algorithm are tabulated.

| Features | Classifiers | Accuracy | Precision | Recall | F-score |
|---|---|---|---|---|---|
| 10000 | Decision tree | 64.2 | 64.2 | 64.2 | 64.2 |
|  | Adaboost | 65.2 | 65.5 | 65.2 | 64.8 |
|  | Randomforest tree | 72.7 | 72.7 | 72.7 | 72.7 |
|  | SVM Linear | 75.1 | 75 | 75.1 | 75 |
|  | SVM rbf | 47.6 | 22.6 | 47.6 | 30.7 |
|  | **Logistics regression** | **75.3** | **75.3** | **75.3** | **75.2** |
| 20000 | Decision tree | 64.2 | 64.2 | 64.3 | 64.3 |
|  | Adaboost | 65 | 65 | 65 | 64.8 |
|  | Randomforest tree | 72.7 | 72.7 | 72.7 | 72.7 |
|  | SVM Linear | 75.5 | 75.5 | 75.5 | 75.5 |
|  | SVM rbf | 47.6 | 22.6 | 47.6 | 30.7 |
|  | **Logistics regression** | **75.6** | **75.6** | **75.6** | **75.6** |
| 30000 | Decision tree | 64.7 | 64.6 | 64.7 | 64.7 |
|  | Adaboost | 65.7 | 65.7 | 65.7 | 65.5 |
|  | Randomforest tree | 72.8 | 72.8 | 72.8 | 72.8 |
|  | **SVM Linear** | **75.7** | **75.7** | **75.7** | **75.7** |
|  | SVM rbf | 47.6 | 22.6 | 47.6 | 30.7 |
|  | Logistics regression | 75.6 | 75.6 | 75.6 | 75.5 |
| 40000 | Decision tree | 64.7 | 64.7 | 64.7 | 64.7 |
|  | Adaboost | 65.3 | 65.3 | 65.3 | 65.1 |
|  | Randomforest tree | 72.9 | 72.8 | 72.9 | 72.8 |
|  | **SVM Linear** | **75.8** | **75.8** | **75.8** | **75.7** |
|  | SVM rbf | 47.6 | 22.6 | 47.6 | 30.7 |
|  | Logistics regression | 75.7 | 75.7 | 75.7 | 75.6 |

Table 2 Results obtained from Classical Machine Learning

| Algorithm | Accuracy | Precision | Recall | F-score | Time for computing |
|---|---|---|---|---|---|
| CNN | **45.25** | 45.25 | 44.61 | 44.68 | 1440 minutes |

Table 3 Results obtained from Deep Learning

## 6 Conclusion

The dataset used in this paper uses the text representation method TFIDF and then given to the classifiers using scikit learn library for classification resulting in getting a state of art accuracy or a benchmark accuracy for this dataset were in the classifier logistics regression out performs well giving a result of about 75.6%. Following the process the dataset is given to the keras embedding technique which extracts the features in the word level is passed to the deep learning algorithm CNN model consisting of a single layer which results in giving an accuracy of about 45.2%. Further this works can be extended by tuning the network following by which good results can be obtained.

**References**

[1] Ortony, A., Clore, G. L., & Collins, A. (1990). The cognitive structure of emotions. Cambridge university press.

[2] Garett, B. (2009). Brain and behaviour: an introduction to biopsychology.

[3] Mohammad, S. M., & Bravo-Marquez, F. (2017). WASSA-2017 shared task on emotion intensity. arXiv preprint arXiv:1708.03700.

[4] Agrawal, A., & An, A. (2012, December). Unsupervised emotion detection from text using semantic and syntactic relations. In Proceedings of the The 2012 IEEE/WIC/ACM International Joint Conferences on Web Intelligence and Intelligent Agent Technology-Volume 01 (pp. 346-353). IEEE Computer Society.

[5] Aizawa, A. (2003). An information-theoretic perspective of tf–idf measures. Information Processing & Management, 39(1), 45-65.

[6] Baziotis, C., Pelekis, N., & Doulkeridis, C. (2017). Datastories at semeval-2017 task 4: Deep lstm with attention for message-level and topic-based sentiment analysis. In Proceedings of the 11th International Workshop on Semantic Evaluation (SemEval-2017) (pp. 747-754).

[7] Peng, C. Y. J., Lee, K. L., & Ingersoll, G. M. (2002). An introduction to logistic regression analysis and reporting. The journal of educational research, 96(1), 3-14.

[8] Bredin, H. (2017, March). Tristounet: triplet loss for speaker turn embedding. In 2017 IEEE International Conference on Acoustics, Speech and Signal Processing (ICASSP) (pp. 5430-5434). IEEE.

[9] Jing, L. P., Huang, H. K., & Shi, H. B. (2002). Improved feature selection approach TFIDF in text mining. In Machine Learning and Cybernetics, 2002. Proceedings. 2002 International Conference on (Vol. 2, pp. 944-946). IEEE.

[10] Mohammad, S., Bravo-Marquez, F., Salameh, M., & Kiritchenko, S. (2018). Semeval-2018 task 1: Affect in tweets. In Proceedings of The 12th International Workshop on Semantic Evaluation (pp. 1-17).

[11] Hocking, R. R. (1976). A Biometrics invited paper. The analysis and selection of variables in linear regression. Biometrics, 32(1), 1-49.

[12] Kim, Y. (2014). Convolutional neural networks for sentence classification. arXiv preprint arXiv:1408.5882.

[13] Vinayakumar, R., Soman, K. P., & Poornachandran, P. (2018). Detecting malicious domain names using deep learning approaches at scale. Journal of Intelligent & Fuzzy Systems, 34(3), 1355-1367.

[14] Vinayakumar, R., Poornachandran, P., & Soman, K. P. (2018). Scalable Framework for Cyber Threat Situational Awareness Based on Domain Name Systems Data Analysis. In Big Data in Engineering Applications (pp. 113-142). Springer, Singapore.

[15] Vinayakumar, R., Soman, K. P., & Poornachandran, P. (2017, September). Deep encrypted text categorization. In Advances in Computing, Communications and Informatics (ICACCI), 2017 International Conference on (pp. 364-370). IEEE.

[16] Vinayakumar, R., Soman, K. P., & Poornachandran, P. (2017, September). Deep android malware detection and classification. In Advances in Computing, Communications and Informatics (ICACCI), 2017 International Conference on (pp. 1677-1683). IEEE.

[17] Mohan, V. S., Vinayakumar, R., Soman, K. P., & Poornachandran, P. (2018, May). Spoof net: Syntactic patterns for identification of ominous online factors. In 2018 IEEE Security and Privacy Workshops (SPW) (pp. 258-263). IEEE.

[18] Vinayakumar, R., Soman, K. P., Poornachandran, P., Mohan, V. S., & Kumar, A. D. (2019). ScaleNet: Scalable and Hybrid Framework for Cyber Threat Situational Awareness Based on DNS, URL, and Email Data Analysis. Journal of Cyber Security and Mobility, 8(2), 189-240.

[19] Vinayakumar, R., Kumar, S. S., Premjith, B., Poornachandran, P., & Padannayil, S. K. (2017). Deep Stance and Gender Detection in Tweets on Catalan Independence@ Ibereval 2017. In IberEval@ SEPLN (pp. 222-229).

[20] Ra, V., HBa, B. G., Ma, A. K., KPa, S., & Poornachandran, P. DeepAnti-PhishNet: Applying Deep Neural Networks for Phishing Email Detection.

[21] Vinayakumar, R., Kumar, S., Premjith, B., Prabaharan, P., & Soman, K. P. DEFT 2017-Texts Search@ TALN/RECITAL 2017: Deep Analysis of Opinion and Figurative language on Tweets in French. In 24e Conférence sur le Traitement Automatique des Langues Naturelles (TALN) (p. 99).

[22] Vinayakumar, R., & Poornachandran, P. (2017). deepCybErNet at EmoInt-2017: Deep Emotion Intensities in Tweets. In Proceedings of the 8th Workshop on Computational Approaches to Subjectivity, Sentiment and Social Media Analysis (pp. 259-263).

[23] Vinayakumar, R., Soman, K. P., & Poornachandran, P. (2017, September). Evaluating shallow and deep networks for secure shell (ssh) traffic analysis. In Advances in Computing, Communications and Informatics (ICACCI), 2017 International Conference on (pp. 266-274). IEEE.

[24] Vinayakumar, R., & Soman, K. P. (2018). DeepMalNet: Evaluating shallow and deep networks for static PE malware detection. ICT Express.

[25] Vazhayil, A., & KP, S. (2018). DeepProteomics: Protein family classification using Shallow and Deep Networks. arXiv preprint arXiv:1809.04461.

[26] HB, B. G., Poornachandran, P., & KP, S. (2018). Deep-Net: Deep Neural Network for Cyber Security Use Cases. arXiv preprint arXiv:1812.03519.

[27] Vinayakumar, R., Soman, K. P., & Naveenkumar, K. S. (2018). Protein Family Classification using Deep Learning. bioRxiv, 414128.